\documentclass[letterpaper, 10 pt, conference]{ieeeconf}  

\IEEEoverridecommandlockouts                              

\usepackage{graphics} 
\usepackage{epsfig} 
\usepackage{mathptmx} 
\usepackage{times} 
\usepackage{amsmath} 
\usepackage{amssymb,bm}  
\usepackage{color}

\usepackage[style=ieee]{biblatex}

\DeclareSourcemap{
  \maps{
    \map{
      \pertype{article}
      \step[fieldset=url, null]
      \step[fieldset=doi, null]
      \step[fieldset=issn, null]
      \step[fieldset=isbn, null]
      \step[fieldset=note, null]
      \step[fieldset=editor, null]
      \step[fieldset=urldate, null]
      \step[fieldset=file, null]
    }
  }
}
\DeclareSourcemap{
  \maps{
    \map{
      \pertype{inproceedings}
      \step[fieldset=url, null]
      \step[fieldset=doi, null]
      \step[fieldset=issn, null]
      \step[fieldset=isbn, null]
      \step[fieldset=note, null]
      \step[fieldset=editor, null]
      \step[fieldset=urldate, null]
      \step[fieldset=file, null]
    }
  }
}
\DeclareSourcemap{
  \maps{
    \map{
      \pertype{incollection}
      \step[fieldset=url, null]
      \step[fieldset=doi, null]
      \step[fieldset=issn, null]
      \step[fieldset=isbn, null]
      \step[fieldset=note, null]
      \step[fieldset=editor, null]
      \step[fieldset=urldate, null]
      \step[fieldset=file, null]
    }
  }
}

\title{\LARGE \bf
Loco-Manipulation with Nonimpulsive \\Contact-Implicit Planning in a Slithering Robot
}

\author{Adarsh Salagame$^{1\text{\textdagger}}$, Kruthika Gangaraju$^{1\text{\textdagger}}$, Harin Kumar Nallaguntla$^{1}$,\\ Eric Sihite$^{2}$, Gunar Schirner$^{1}$, Alireza Ramezani$^{1*}$
\thanks{$^{1}$This author is with the Department of Electrical and Computer Engineering, Northeastern University, Boston MA
        {\tt\small salagame.a, gangaraju.k, nallaguntla.h, G.Schirner, a.ramezani@northeastern.edu*}}%
\thanks{$^{2}$ This author is with California Institute of Technology, Pasadena CA
		{\tt\small esihite@caltech.edu}}%
\thanks{\text{\textdagger}These authors have equal contribution to this work.}
\thanks{$*$Indicates the corresponding author.}
}

\begin{document}

\maketitle
\thispagestyle{empty}
\pagestyle{empty}

\begin{abstract}

Object manipulation has been extensively studied in the context of fixed base and mobile manipulators. However, the overactuated locomotion modality employed by snake robots allows for a unique blend of object manipulation through locomotion, referred to as loco-manipulation. The following work presents an optimization approach to solving the loco-manipulation problem based on non-impulsive implicit contact path planning for our snake robot COBRA. We present the mathematical framework and show high-fidelity simulation results and experiments to demonstrate the effectiveness of our approach.

\end{abstract}

\section{Introduction}
\label{sec:intro}

Optimization-driven path planning and control strategies have emerged as pivotal methodologies for managing diverse contact-intensive systems within real-world experimental settings \cite{sihite_multi-modal_2023,sihite_unilateral_2021,sihite_efficient_2022,dangol_towards_2020,dangol_performance_2020,liang_rough-terrain_2021,dangol_reduced-order-model-based_2021}. These approaches have found widespread application across various locomotion modalities, such as legged and slithering locomotion, showcasing remarkable efficacy, including rapid contact planning in terrestrial environments \cite{ding_real-time_2019,carius_trajectory_2018, droge_optimal_2012, chang_optimal_2018, hicks_method_2005, mohammadi_optimal_2007}. Notably, point-contact models such as legged robots \cite{grizzle_progress_nodate,dangol_control_2021,ramezani_generative_2021,salagame_quadrupedal_2023,salagame_letter_2022} have been particularly receptive to optimization techniques compared to systems characterized by extensive contact interactions, such as snake robots. Given the intricate dynamics inherent in slithering systems, which encompass sophisticated contact dynamics \cite{fu_robotic_2020,chen_studies_2004}, there arises a pressing need for enhanced modeling and control methodologies. These tools are indispensable for orchestrating body movements through the modulation of joint torques, ground reaction forces, and the coordination of contact sequences comprising timing and spatial positioning.

Snake locomotion embodies a spectrum of techniques tailored to diverse environments and challenges. Lateral undulation, as exemplified in studies by \cite{wiriyacharoensunthorn_analysis_2002, ma_analysis_2003, ma_analysis_1999}, relies on anisotropic friction to propel snakes forward in a sinusoidal trajectory. Rectilinear motion, elucidated in works such as \cite{rincon_ver-vite_2003, ohno_design_2001}, involves the controlled compression and expansion of scales to facilitate longitudinal movement, ideal for maneuvering through constrained spaces. The sidewinding gait, as demonstrated in investigations like \cite{liljeback_modular_2005, burdick_sidewinding_1994}, is employed on slippery or sandy terrains, featuring a sinusoidal motion for lateral displacement. In confined settings, snakes adopt the concertina gait, outlined in \cite{shan_design_1993}, characterized by coiling and uncoiling actions to progress longitudinally. Additionally, unconventional gaits such as the inchworm, slinky, lateral rolling, and tumbling locomotion, proposed in studies like \cite{yim_new_1994, rincon_ver-vite_2003}, exploit the articulated structure of the snake's body to manifest unique locomotive patterns.

While the predominant focus of snake robotics research lies in emulating snake locomotion and replicating its distinctive movement patterns using central pattern generators, optimization remains underutilized. Leveraging optimization techniques can propel snake robots beyond the mentioned locomotion feats above, enabling the exploration of alternative locomotion modes achievable through optimal joint movements that respect complementarity conditions. These optimization tools unlock the inherent redundancy and intricate articulation of snake bodies for acyclic locomotion and manipulation (loco-manipulation), which have often been overlooked in favor of standard locomotion feats.

This study endeavors to investigate the loco-manipulation capabilities facilitated by the redundant body structures of a snake robot called COBRA, particularly concerning its interactions with objects. Addressing this contact-intensive challenge presents intriguing prospects for harnessing contact-implicit optimization, a prevalent design paradigm within locomotion research that, inexplicably, remains relatively unexplored within the domain of snake-type robots.

\begin{figure}
    \centering
    \includegraphics[width=1.0\linewidth]{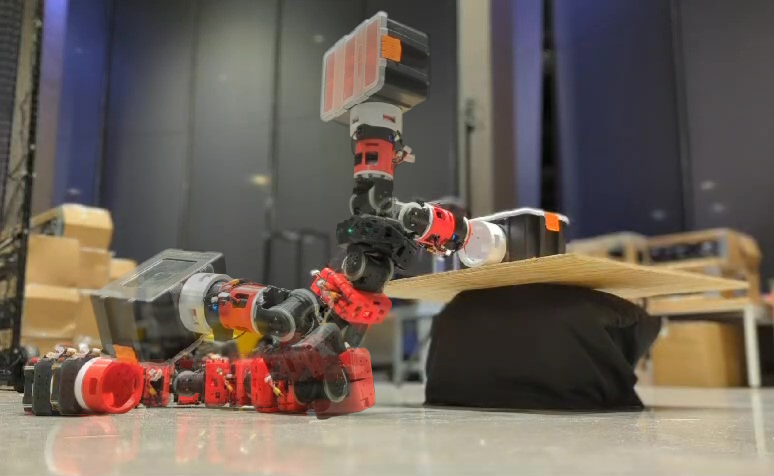}
    \caption{Illustrates COBRA while manipulating the position of a box.}
    \label{fig:cover-image}
\end{figure}

\subsection{COBRA Platform}

As illustrated in Fig.~\ref{fig:cover-image}, the COBRA system \cite{jiang_snake_2023,jiang_hierarchical_2023,salagame_how_2023} is comprised of eleven actuated joints. Positioned at the frontal segment is a head module housing the onboard computing system, a communication module, and an inertial measurement unit (IMU) utilized for navigation. The rear section houses an interchangeable payload module, accommodating additional electronics tailored to specific tasks undertaken by COBRA. The remaining components consist of identical modules, each housing a joint actuator and a battery.

\begin{figure}
    \centering
    \includegraphics[width=1\linewidth]{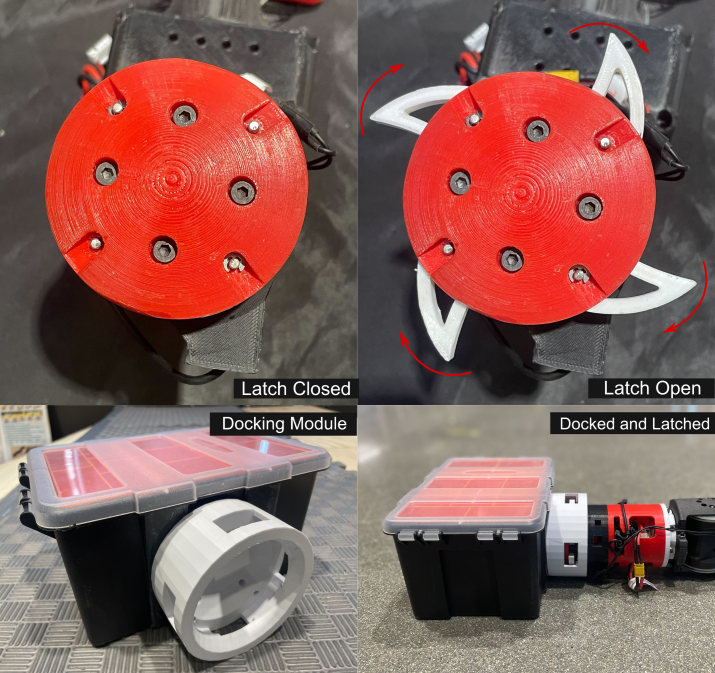}
    \caption{(Above) Closeup view of the head with actuated fins. (Below) Docking module attached to object for loco-manipulation.}
    \label{fig:head}
\end{figure}

A latching mechanism (see Fig.~\ref{fig:head}) integrated into the head module allows for the attachment of a gripper, enabling object manipulation through conventional grip-based methods. This mechanism features a Dynamixel XC330 actuator situated within the head module, driving a central gear. The gear interfaces with partially geared sections of four fin-shaped latching fins. Each latching fin's curved outer face spans an arc length equal to 1/4 of the circumference of the head module's circular cross-section. When retracted, these four fins form a thin cylinder aligned with the cylindrical face of the head module. A dome-shaped cap positioned at the end of the head module houses the fins between it and the main body of the head module, with clevis pins utilized to secure the fins in place.

\subsection{Contributions and Paper Organization}

The primary research objectives encompass: Investigating optimal control design approaches to effectively guide the joints along desired trajectories for object manipulation. This work introduces an optimization approach based on non-impulsive contact-implicit path planning for COBRA. We demonstrate the effectiveness of this method in generating optimal joint trajectories for desired object movements across flat and ramp surfaces in simulation and experiment.

The paper's structure is as follows: 
We begin by introducing the fundamental concept underlying the motion optimization approach employed in this study, elucidating the incorporation of contact forces within the context of object locomotion and manipulation. We then present results in simulation and in experiment demonstrating object manipulation in various scenarios. Finally, we discuss paths for future development. 

\section{Nonimpulsive Contact-implicit Motion Optimization}
\label{sec:mo-opt}

\begin{figure}
    \centering
    \includegraphics[width=1\linewidth]{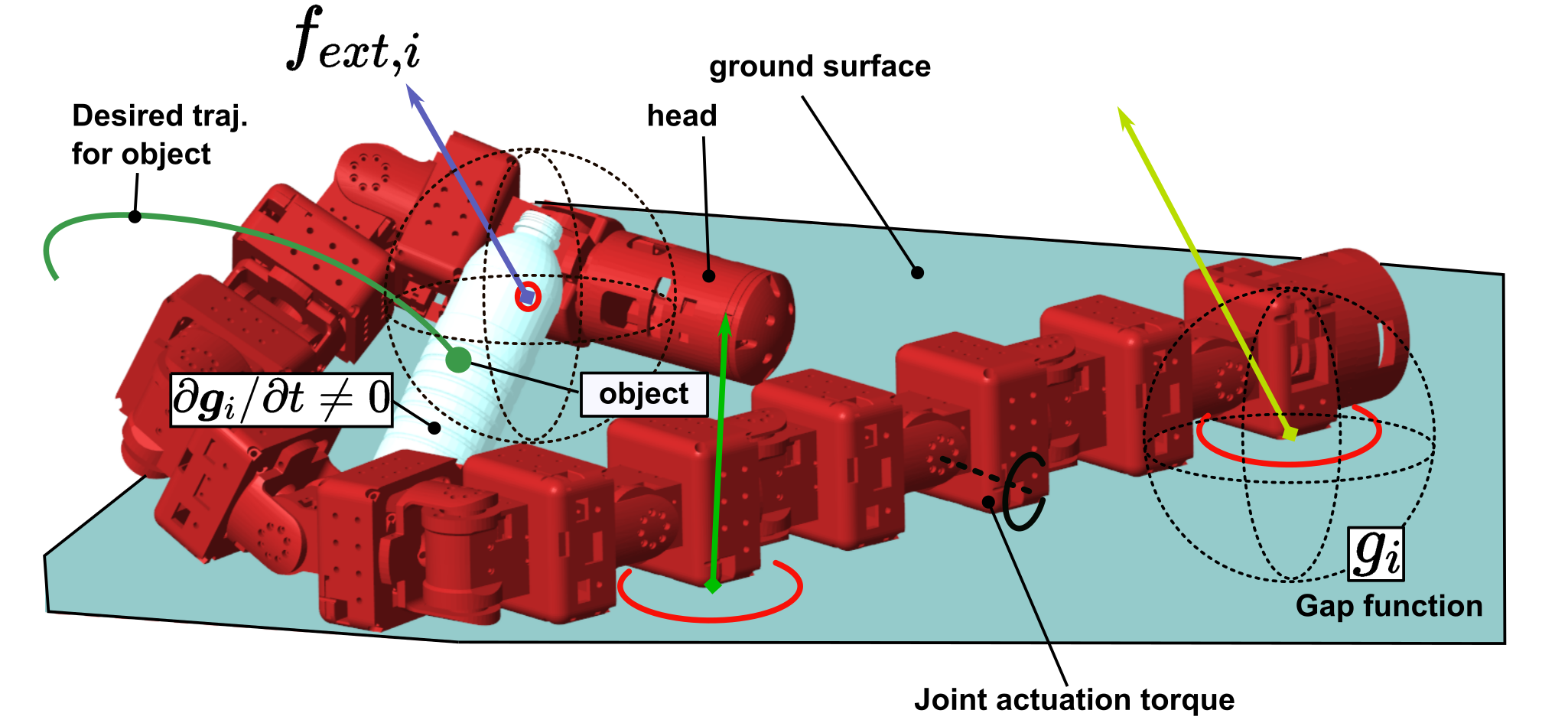}
    \caption{Full-dynamics model parameters in the object manipulation task considered in this paper}
    \label{fig:full-dyn}
\end{figure}

The motion dynamics governing the COBRA snake robot, equipped with 11 body joints, are succinctly captured in the following equations of motion:
\begin{equation}
    \begin{aligned}
        \bm M(\bm q) \dot{\bm u} - \bm h(\bm q, \bm u, \bm \tau) &= \sum_i \bm J_i^{\top}(\bm q) \bm f_{ext,i},\\
        \bm h(\bm q, \bm u, \bm \tau) &= \bm C(\bm q,\bm u)\bm u + \bm G(\bm q) + \bm B(\bm q)\bm \tau
    \end{aligned}
    \label{eq:eom}
\end{equation}
In this expression, the mass-inertia matrix $\bm M$ operates in a space of dimensionality $\mathbb{R}^{17 \times 17}$, the terms encompassing centrifugal, Coriolis, gravity, and actuation ($\bm \tau$) are succinctly represented by $\bm h \in \mathbb{R}^{17}$, the external forces $\bm f_{ext,i}$ and their respective Jacobians $\bm J_i$ reside in the space $\mathbb{R}^{3 \times 17}$. For clarity and conciseness, specifics regarding the generalized coordinates and velocities of COBRA are omitted.

In the object manipulation problem depicted in Fig.~\ref{fig:full-dyn}, the external forces stem solely from active unilateral constraints, such as contact forces between the ground surface and the robot or between a movable object and the robot. This assumption conveniently establishes a complementary relationship, where the product of two variables, including force and displacement, in the presence of holonomic constraints is zero, between the separation $\bm{g}_i$ (the gap between the body, terrain, and object) and the force exerted by a hard unilateral contact.

The concept of normal cone inclusion on the displacement, velocity, and acceleration levels from \cite{studer_numerics_2009} permits the expression:
\begin{equation}
    \begin{aligned}
        -\bm g_i & \in \partial \Psi_{i}\left(\bm f_{ext,i}\right) \equiv \mathcal{N}_{\mathcal{F}_i}\left(\bm f_{ext,i}\right) \\
        -\dot{\bm g}_i & \in \partial \Psi_{i}\left(\bm f_{ext,i}\right) \equiv \mathcal{N}_{\mathcal{F}_i}\left(\bm f_{ext,i}\right) \\
        -\ddot{\bm g}_i & \in \partial \Psi_{i}\left(\bm f_{ext,i}\right) \equiv \mathcal{N}_{\mathcal{F}_i}\left(\bm f_{ext,i}\right)
    \end{aligned}
    \label{eq:normal-cone-inclusion}
\end{equation}
where $\Psi_i(.)$ denotes the indicator function. The gap function $\bm{g}_i$ is defined such that its total time derivative yields the relative constraint velocity $\dot{\bm{g}}_i = \bm W_i^{\top} \bm u + \zeta_i$, where $\bm W_i = \bm W_i(\bm q, t) = \left( \partial \bm{g}_i / \partial \bm q \right)^{\top}$ and $\zeta_i = \zeta_i(\bm q, t) = \partial \bm{g}_i / \partial t$. 

In the context of the primary objectives of loco-manipulation with COBRA, various conditions of the normal cone inclusion as described in Eq.~\ref{eq:normal-cone-inclusion} were explored. In scenarios where nonimpulsive unilateral contact forces are employed to manipulate rigid objects (e.g., the box shown in Fig.~\ref{fig:full-dyn}), $\partial \bm{g}_i / \partial t \neq 0$. This factor holds significant importance in motion planning considered in this work and is enforced during optimization.

\begin{figure*}
    \centering
    \includegraphics[width=1\linewidth]{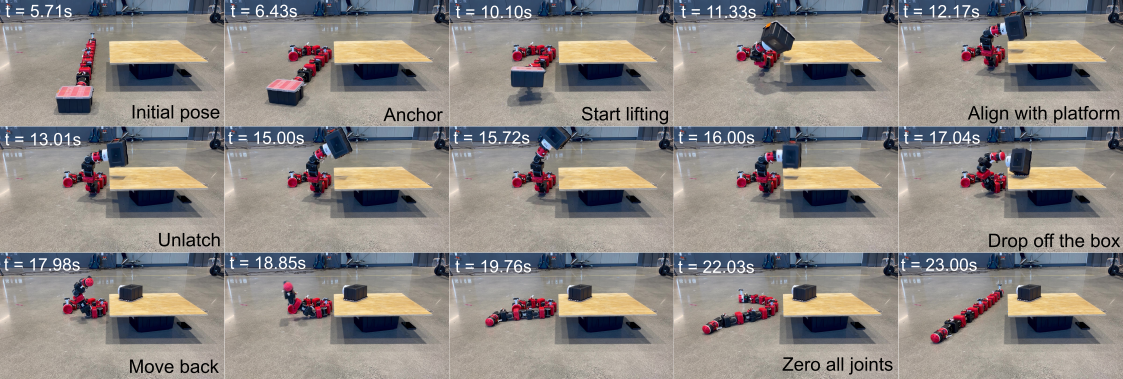}
    \caption{Snapshots of COBRA lifting a box and placing it on a raised platform.}
    \label{fig:liftnplace}
\end{figure*} 

\begin{figure*}
    \centering
    \includegraphics[width=1\linewidth]{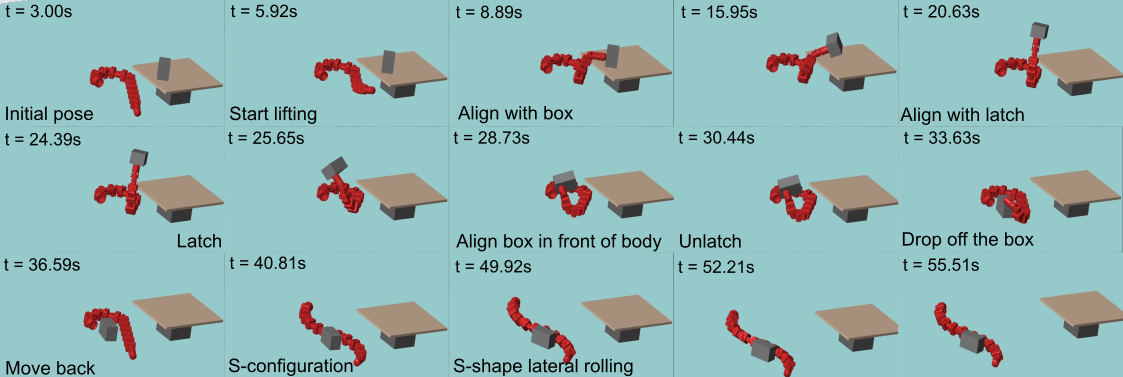}
    \caption{Snapshots depict simulation of COBRA lifting a box from a raised platform, placing it on the flat ground, and translating the box to a new location through continuous body-object interactions during slithering motions.}
    \label{fig:picknplace-sim}
\end{figure*}

\begin{figure*}
    \centering    \includegraphics[width=1\linewidth]{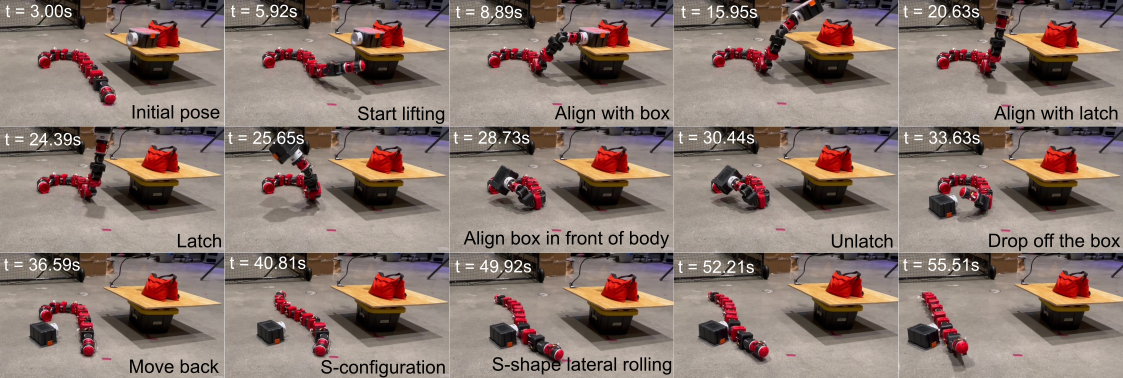}
    \caption{Snapshots depict COBRA lifting a box from a raised platform, placing it on the flat ground, and translating the box to a new location through continuous body-object interactions during slithering motions.}
    \label{fig:picknplace}
\end{figure*}

The total time derivative of the relative constraint velocity yields the relative constraint accelerations $\ddot{\bm g}_i=\bm W^{\top} \dot{\bm u}+\hat{\zeta}_i$ where $\hat{\zeta}_i=\hat{\zeta}_i(\bm q, \bm u, t)$. We describe a geometric constraint on the acceleration level such that the initial conditions are fulfilled on velocity and displacement levels:
\begin{equation}
    \begin{aligned}
        &\bm g_i(\bm q, t)=0, \\
        &\dot{\bm g}_i=\bm W_i^\top \bm u+\zeta_i=0, \\
        &\ddot{\bm g}_i=\bm W_i^\top \dot{\bm u}+\hat \zeta_i=0,\\
        &\dot{\bm g}_i\left(\bm q_0, \bm u_0, t_0\right)=0,\\
        &\partial \bm{g}_i / \partial t \neq 0
    \end{aligned}   
\end{equation}
which implies that the generalized constraint forces must be perpendicular to the manifolds $\bm g_i=0$, $\dot{\bm g_i}=0$, and $\ddot{\bm g}_i=0$. This formulation directly accommodates the integration of friction laws, which naturally pertain to velocity considerations. We divide the contact forces into normal and tangential components, denoted as $\bm f_{ext,i}=\left[\begin{array}{ll}f_{N,i}, & \bm f_{T,i}^{\top}\end{array}\right]^{\top} \in \mathcal{F}_{i}$. 

In this context, the force space $\mathcal{F}_i$ facilitates the specification of non-negative normal forces ($\mathbb{R}_{0}^+$) and tangential forces adhering to Coulomb friction $\left\{\bm f_{T,i} \in \mathbb{R}^2, ||\bm f_{T,i}||<\mu\left|f_{N,i}\right| \right\}$, with $\mu$ representing the friction coefficient.

The underlying rationale behind this approach is that while the force remains confined within the interior of its designated subspace, the contact velocity remains constrained to zero. Conversely, non-zero gap velocities only arise when the forces reach the boundary of their permissible set, indicating either a zero normal force or the maximum friction force opposing the direction of motion.

\begin{figure*}
    \centering
    \includegraphics[width=1\linewidth]{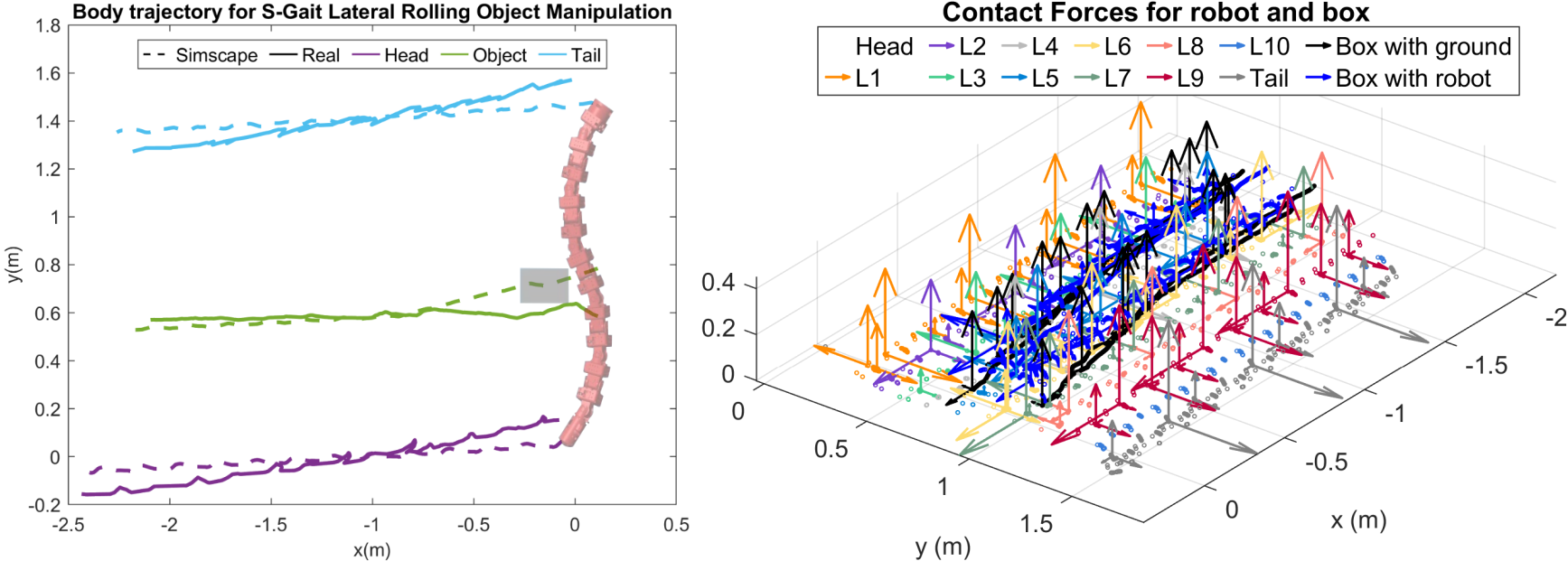}
    \caption{(Left) Comparison of the robot and object trajectory with simulation for S-shape lateral rolling gait on flat ground shown from a top view. (Right) Contact forces for robot-ground, robot-box, and box-ground contacts.}
    \label{fig:s-lateral-roll}
\end{figure*}

To proceed with the loco-manipulation problem considered here, it proves advantageous to reconfigure Eq.~\ref{eq:eom} into local contact coordinates (task space). This can be achieved by recognizing the relationship:
\begin{equation}
    \dot{\bm g}=\bm J_{\mathrm{c}} \bm u,     
\end{equation}
where $\bm g$ and $\bm J_c$ represent the stacked contact separations and Jacobians, respectively. By differentiating the above equation with respect to time and substituting Eq.~\ref{eq:eom}, we obtain:
\begin{equation}
    \ddot{\bm g} = \bm J_c \bm M^{-1} \bm J_c^{\top} \bm f_{ext} + \dot{\bm J}_c \bm u + \bm J_c \bm M^{-1} \bm h,
\end{equation} 
where $\bm G = \bm J_c \bm M^{-1} \bm J_c^{\top}$ -- the Delassus matrix -- signifies the apparent inverse inertia at the contact points, and $ \bm c = \dot{\bm J}_c \bm u + \bm J_c \bm M^{-1} \bm h$ encapsulates all terms independent of the stacked external forces $\bm f_{ext}$. At this point, the principle of least action asserts that the contact forces are determined by the solution of the constrained optimization problem:
\begin{equation}
    \begin{aligned}
        &\underset{\left\{\bm f_{ext,i},\bm u\right\}}{\operatorname{\textbf{minimize}}}~\frac{1}{2} \bm f_{ext}^\top \bm G \bm f_{ext}+\bm f_{ext}^\top \bm c \\
        &{\operatorname{\textbf{s.t.}}}\\
        &{\operatorname{\textbf{(1)}}~\bm M(\bm q) \dot{\bm u} - \bm h(\bm q, \bm u, \bm \tau) - \sum_i \bm J_i^{\top}(\bm q) \bm f_{ext,i}}=0\\
        &{\operatorname{\textbf{(2)}}}~\|\bm q\|\leq q_{max}\\
        &{\operatorname{\textbf{(3)}}}~\|\bm \tau\|\leq \tau_{max}\\
    \end{aligned}
\end{equation}
where $q_{max}$ and $\tau_{max}$ denote maximum joint movements and actuation torques, respectively. In the above optimization problem, (1), (2), and (3) enforce dynamics agreement, kinematics restrictions, and actuation saturations, respectively. 

Subsequently, a time-stepping methodology facilitates the integration of system dynamics across a time interval $\Delta t$ while internally addressing the resolution of contact forces. We employ the shooting method to find the optimal joint positions $\bm u_{ref}$ for minimizing $\frac{1}{2} \bm f_{ext}^\top \bm G \bm f_{ext}+\bm f_{ext}^\top \bm c$, ensuring that the generalized contact forces $\bm f_{ext}$ are orthogonal to gap functions and their derivatives.

\section{Results}
\label{sec:results}
We demonstrate the object manipulation with COBRA by implementing the obtained desired trajectories in an open-loop fashion to show the effectiveness of our approach in preparation for the hardware implementation of the presented optimization model in closed-loop fashion. 

We have developed a high-fidelity simulation using Matlab Simulink that enables studying parameters such as ground contact forces and object interactions that are difficult to obtain ground truth for on the real robot. Figure~\ref{fig:liftnplace} shows COBRA lifting an object from the ground and placing it on a raised platform of height $0.3$m. The robot starts with the object docked to the head module using the docking module shown in Fig. \ref{fig:head}. It curls the tail to increase the region of support, allowing it to stay balanced while lifting the box up to a height of $0.4$m from the ground. Once it has moved the box over the platform, it unlatches from the object and shakes its head to dislodge it and deposit it on the platform. 

In Fig. \ref{fig:picknplace-sim} and Fig.~\ref{fig:picknplace}, first in simulation, then in experiment, the robot starts on the ground with the object on the raised platform. The robot is able to, once again, curl its tail and lift the head to align itself with the docking module on the object. With some manipulation to allow the box to align to the latch, the robot can latch onto the box and place it in front of the body. From this position, COBRA can use its lateral rolling gait to move the object to a desired location. 

\begin{figure*}
    \centering
    \includegraphics[width=1\linewidth]{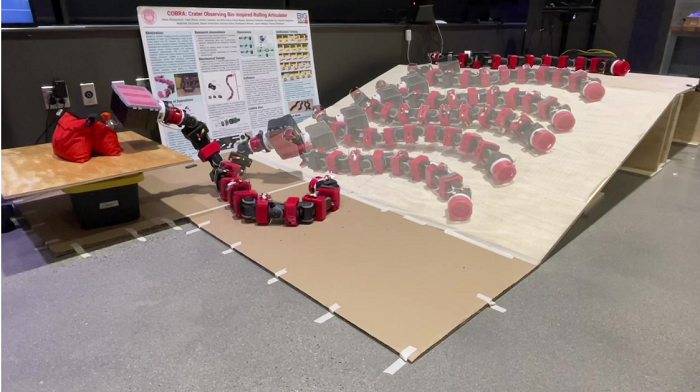}
    \caption{Snapshots capture COBRA lifting a box from a raised platform, setting it down on the ground, and ascending a ramp by continually pushing the box while lateral rolling.}
    \label{fig:ramp}
\end{figure*}

Figure \ref{fig:s-lateral-roll} shows the body trajectory and the box trajectory for the S-shape lateral rolling gait employed by the robot to move the box on flat ground, along with the associated ground contact forces obtained from simulation. Extending this further, Fig.~\ref{fig:ramp} shows COBRA picking up the box from a raised platform, placing it in front of the body and moving the box to the top of a ramp.

\section{Concluding Remarks}
\label{sec:conclusion}

This paper presents an optimization-based approach to path planning and object manipulation for loco-manipulation, employing COBRA, a morpho-functional robot equipped with manifold moving joints and contact-rich behavior. With 11 actuated joints and onboard sensing and computation capabilities, COBRA enables high-fidelity modeling for manipulating objects on flat surfaces, utilizing nonimpulsive contact-implicit path planning. Simulation and real-world experiments demonstrate successful object manipulation and locomotion, supported by an analysis of ground reaction forces and intermittent contact locations. In future research, we aim to enhance COBRA's capabilities by integrating tactile sensors for online and real-time path planning/re-planning based on feedback from the environment.


\begin{figure}
    \centering
    \includegraphics[width=1\linewidth]{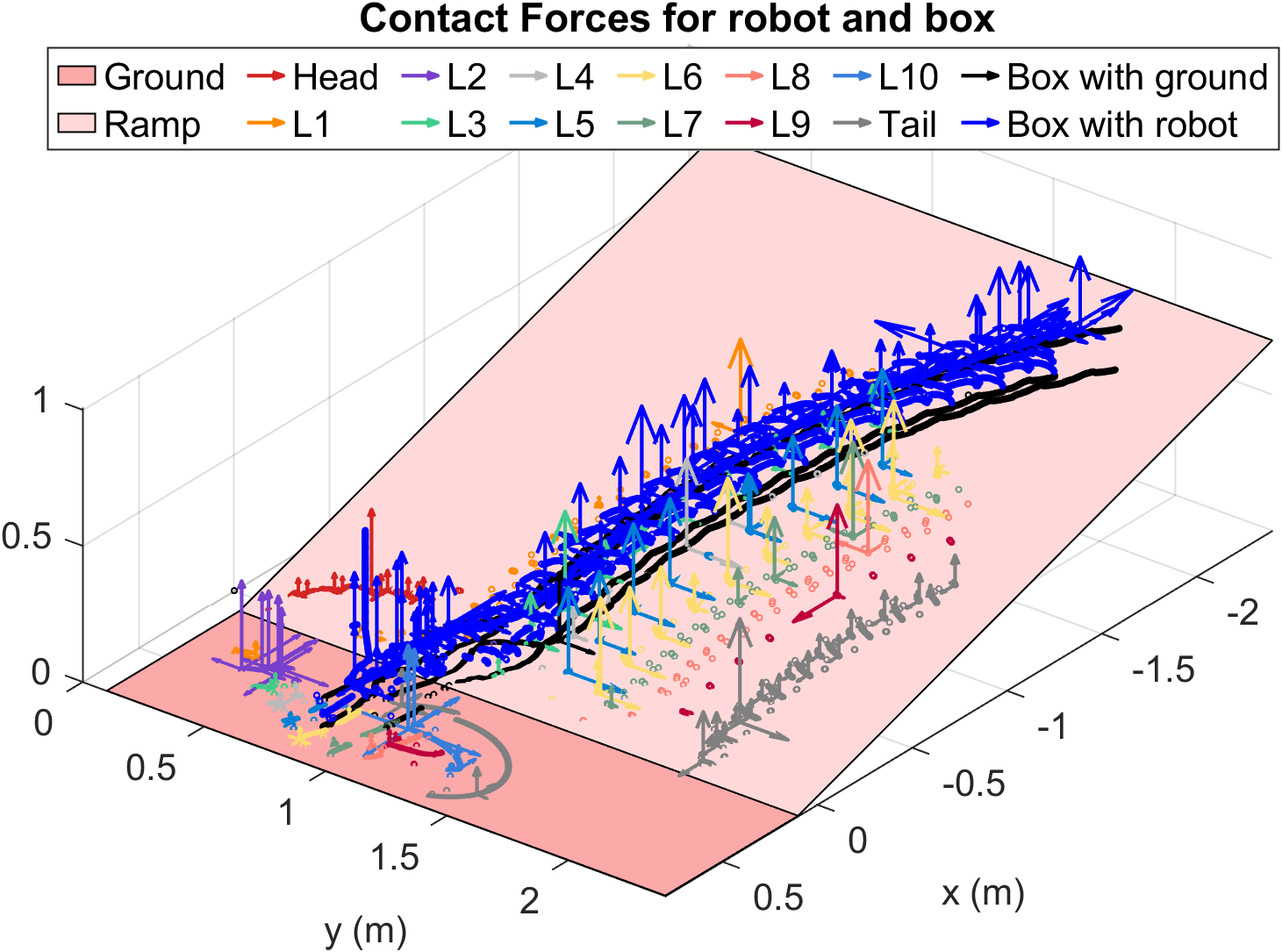}
    \caption{Contact forces for the robot and box during picking the box from a raised platform, placing on the ground, and manipulation using s-lateral rolling gait up a ramp.}
    \label{fig:enter-label}
\end{figure}


\nocite{bayraktaroglu_understanding_2004, lu_3d_2006, takemori_gait_2018, reyes_studying_2017, lee_efficient_2019, rakhimkul_autonomous_2019, okada_object-handling_1979, qiu_reinforcement_2021, taal_3_2009, ramesh_sensnake_2022, nansai_dynamic_2016, reyes_snake_2017,holden_optimal_2014}
\printbibliography

\end{document}